\documentclass[default,iicol]{sn-jnl}

\usepackage{graphicx}%
\usepackage{multirow}%
\usepackage{amsmath,amssymb,amsfonts}%
\usepackage{amsthm}%
\usepackage{mathrsfs}%
\usepackage[title]{appendix}%
\usepackage{xcolor}%
\usepackage{textcomp}%
\usepackage{manyfoot}%
\usepackage{booktabs}%
\usepackage{algorithm}%
\usepackage{algorithmicx}%
\usepackage{algpseudocode}%
\usepackage{listings}%
\usepackage{bm}
\usepackage{eqnarray}
\usepackage{todonotes}[colorinlistoftodos]
\usepackage{bbm}
\usepackage{soul}

\usepackage{hyperref}
\hypersetup{colorlinks,linkcolor={black},citecolor={cyan},urlcolor={magenta}}

\usepackage{algorithm} 
\usepackage{algpseudocode} 
\usepackage{lipsum}
\usepackage{subcaption}

\newcommand{\ra}[1]{\renewcommand{\arraystretch}{#1}}

\usepackage[font=small,labelfont=bf]{caption}

\begin{document}

\title{R$\times$R: Rapid eXploration for Reinforcement Learning via Sampling-based Reset Distributions and Imitation Pre-training}


\author*[1]{\fnm{Gagan} \sur{Khandate}}\email{gagank@cs.columbia.edu}
\equalcont{These authors contributed equally to this work.}

\author*[1]{\fnm{Tristan} \sur{L. Saidi}}\email{tls2160@columbia.edu}
\equalcont{These authors contributed equally to this work.}

\author[1]{\fnm{Siqi} \sur{Shang}}
\equalcont{These authors contributed equally to this work.}

\author[2]{\fnm{Eric T.} \sur{Chang}}
\author[2]{\fnm{Yang} \sur{Liu}}
\author[2]{\fnm{Seth} \sur{Dennis}}
\author[2]{\fnm{Johnson} \sur{Adams}}
\author[2]{\fnm{Matei} \sur{Ciocarlie}}

\affil[1]{\orgdiv{Department of Computer Science}}
\affil[2]{\orgdiv{Department of Mechanical Engineering}}
\affil[]{\orgaddress{Columbia University, \city{New York}, \postcode{10027}, \state{NY}, \country{USA}}}




\abstract{
We present a method for enabling Reinforcement Learning of motor control policies for complex skills such as dexterous manipulation.  We posit that a key difficulty for training such policies is the difficulty of exploring the problem state space, as the accessible and useful regions of this space form a complex structure along manifolds of the original high-dimensional state space. This work presents a method to enable and support exploration with Sampling-based Planning. We use a generally applicable non-holonomic Rapidly-exploring Random Trees algorithm and present multiple methods to use the resulting structure to bootstrap model-free Reinforcement Learning. Our method is effective at learning various challenging dexterous motor control skills of higher difficulty than previously shown. In particular, we achieve dexterous in-hand manipulation of complex objects while simultaneously securing the object without the use of passive support surfaces. These policies also transfer effectively to real robots. A number of example videos can also be found on the project website: \href{sbrl.cs.columbia.edu}{sbrl.cs.columbia.edu}

}

\keywords{Exploration, Reinforcement Learning, Dexterous Manipulation}



\maketitle

\section{Introduction}

Reinforcement Learning (RL) of robot sensorimotor control policies has seen great advances in recent years, demonstrated for a wide range of motor tasks such as locomotion and manipulation. It has enabled impressive locomotion skills such as walking in rough terrain and parkour \cite{Agarwal2022-rp, Zhuang2023-bi}. In the case of manipulation, this has translated into higher levels of dexterity than previously possible, typically demonstrated by the ability to reorient a grasped object in-hand using complex finger movements \citep{OpenAI2019-ng, Chen2021-ig, Qi2022-wy}. 


However, training a sensorimotor policy is still a difficult process, particularly for hard-exploration problems where the underlying state space exhibits complex structure, such as "narrow passages" between parts of the space that are accessible or useful. Manipulation is indeed such a problem: even when starting with the object secured between the digits, a random action can easily lead to a drop, and thus to an irrecoverable state. Finger-gaiting \cite{Ma2011-fo} further implies transitions between different subsets of fingers used to hold the object, all while maintaining stability. This leads to difficulty in exploration during training, since random perturbations in the policy action space are unlikely to discover narrow passages in state space. Current studies address this difficulty through a variety of means: using simple, convex objects and non-convex objects of significantly reduced size to reduce the difficulty of the task, reliance on support surfaces to reduce the chances of a drop, object pose tracking through extrinsic sensing, etc.

The difficulty of exploring problems with labyrinthine state space structures is far from new in robotics. In fact, the large and highly effective family of Sampling-Based Planning (SBP) algorithms was developed to address this exact problem. By expanding a known structure towards targets randomly sampled in the state space of the problem, SBP methods can explore even very high-dimensional state spaces in ways that are probabilistically complete, or guaranteed to converge to optimal trajectories. However, SBP algorithms are traditionally designed to find trajectories rather than policies. For problems with computationally demanding dynamics, SBP cannot be used online for previously unseen start states or to quickly correct when unexpected perturbations are encountered along the way.

\begin{figure*}[t]
    \centering
    \includegraphics[width=\textwidth]{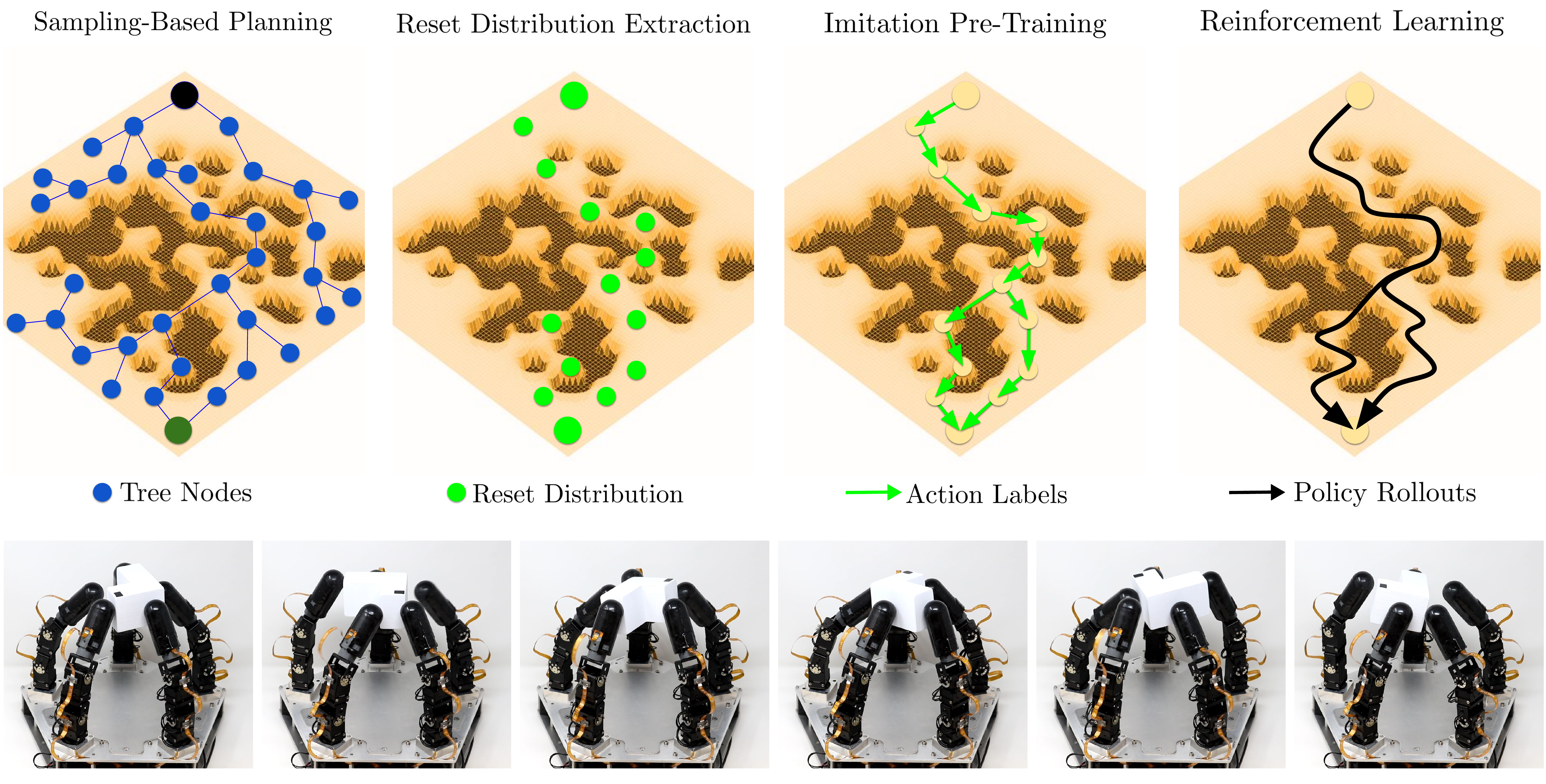}
    \caption{Our method illustrated with an abstract state-space consisting of narrow stable regions (beige) between large unstable regions (black). The proposed two-stage approach uses sampling-based planning to explore the challenging state-space and leverages the information within best paths by using state data as reset distribution and action data for imitation pre-training for efficient reinforcement learning.}
    \label{fig:rxr}
\end{figure*}

Broadly, this work draws on the strength of both RL and SBP methods in order to train motor control policies for in-hand manipulation with finger gaiting. We aim to manipulate more difficult objects, including concave shapes, while securing them at all times without relying on support surfaces. Furthermore, we aim to achieve large reorientation of the grasped object with purely intrinsic (tactile and proprioceptive) sensing. To achieve that we use a non-holonomic RRT algorithm with added constraints to find approximate trajectories that explore the useful parts of the problem state space. Then, we use these trajectories towards training complete RL policies based on the full dynamics of the problem. In particular, we use state data sampled from the tree to construct exploratory reset distribution and use action data from tree paths to obtain a warm-start policy through imitation learning and illustrate this in Fig \ref{fig:rxr}.

This extended version of our study contains a number of additions compared to the previous conference proceedings version~\cite{Khandate2023-fc}. We introduce Imitation Pre-training, a method for leveraging actions sampled from the RRT tree in order to further bootstrap learning and show that it further improves training efficiency. We extend our applications to the more difficult in-hand manipulation task of reorientation to a desired pose, which can be either a canonical pose known at training time (for example, as needed by downstream steps in an assembly process) or arbitrary and specified as a goal at run-time. Finally, we provide a number of additional insights obtained from an extensive ablation of our approach.

Overall, the main contributions of this work include:
\begin{itemize}
    \item To the best of our knowledge, we are the first to show that reset distributions generated via SBP with kinematic constraints can enable more efficient training of RL control policies for dexterous in-hand manipulation. 
    \item We show that SBP can explore useful parts of the manipulation state space, allowing RL to later fill in the gaps between approximate trajectories by learning appropriate actions under more realistic dynamic constraints. A warm-start policy obtained by imitating the actions from sampling-based trajectories provides an additional boost in training speed. 
    \item The exploration boost from SBP allows us to train policies for dexterous skills not previously shown, such as in-hand manipulation of concave shapes, with only intrinsic sensing and no support surfaces. We demonstrate these skills both in simulation and on real hardware.
\end{itemize}

\section{Related Work}
\vspace{3mm}
\noindent \textbf{Exploration in RL}.  Exploration methods for general RL operate under the strict assumption that the learning agent cannot teleport between states, mimicking the constraints of the real world. Under such constraints, proposed exploration methods include using intrinsic rewards \cite{Pathak2017-ik, Haarnoja2018-zj} or improving action consistency via temporally correlated noise in policy actions \cite{Amin2021-dm} or parameter space noise \cite{ Plappert2017-ba}. 

Fortunately, in cases where the policies are primarily trained in simulation, this requirement can be relaxed, and we can use our knowledge of the relevant state space to design effective exploration strategies. A number of these methods improve exploration by injecting useful states into the reset distribution during training. \citet{Nair2017-nu} use states from human demonstrations in a block stacking task, while Ecoffet et al. \cite{Ecoffet2019-lk, Ecoffet2021-xs} use states previously visited by the learning agent itself for problems such as Atari games and robot motion planning. \citet{Tavakoli2018-ah} evaluate various schemes for maintaining and resetting from the buffer of visited states. However, these schemes were evaluated only on benchmark continuous control tasks \cite{Duan2016-em}. From a theoretical perspective, \citet{Agarwal2020-pp} show that a favorable reset state distribution provides a means to circumvent worst-case exploration issues using sample complexity analysis of policy gradients. 

\vspace{3mm}
\noindent \textbf{Combining SBP and RL}. Finding feasible trajectories through a complex state space is a well-studied motion planning problem. Of particular interest to us are sampling-based methods such as Rapidly-exploring Random Trees (RRT) \cite{LaValle1998-kn, Karaman2010-zi, Webb2013-nt} and Probabilistic Road Maps (PRM) \cite{Kavraki1996-gr, Kavraki1998-wl}. These families of methods have proven highly effective and are still being expanded. Stable Sparse-RRT (SST) and its optimal variant SST* \cite{Li2021-lv} are examples of recent sampling-based methods for high-dimensional motion planning with physics. However, the goal of these methods is finding (kinodynamic) trajectories between known start and goal states rather than closed-loop control policies which can handle deviations from the expected states.

Several approaches have tried to combine the exploratory ability of SBP with RL, leveraging planning for global exploration while learning a local control policy via RL \cite{Chiang2019-vm,Francis2020-qe,Schramm2022-hs}. These methods were primarily developed for and tested on navigation tasks, where nearby state space samples are generally easy to connect by an RL agent acting as a local planner. The LeaPER algorithm \cite{Pinto2018-xi} also uses plans obtained by RRT as reset state distribution and learns policies for simple non-prehensile manipulation. However, the state space for the prehensile in-hand manipulation tasks we show here is highly constrained, with small useful regions and non-holonomic transitions. Other approaches use trajectories planned by SBP as expert demonstrations. \citet{Morere2020-gq} recommend using a policy trained with SBP as expert demonstrations as an initial policy. Alternatively, \citet{Jurgenson2019-ye} and \citet{Ha2020-od} use planned trajectories in the replay buffer of an off-policy RL agent for multi-arm motion planning. First, these methods requires that planned trajectories also include the actions used to achieve transitions, which SBP does not always provide. Next, it is unclear how off-policy RL can be combined with the extensive physics parallelism that has been vital in the recent success of on-policy methods for learning manipulation \cite{Allshire2021-qp, Makoviychuk2021-ko, Chen2021-ig}. 

\vspace{3mm}
\noindent \textbf{Dexterous Manipulation}.
Turning specifically to the problem of dexterous manipulation, a number of methods have been used to advance the state of the art, including planning, learning, and leveraging mechanical properties of the manipulator. \citet{Leveroni1996-iy} build a map of valid grasps and use search methods to generate gaits for planar reorientation, while \citet{Han1998-xj} consider finger-gaiting of a sphere and identify the non-holonomic nature of the problem. Some methods have also considered RRT for finger-gaiting in-hand manipulation \cite{Yashima2003-lw, Xu2007-yb}, but limited to simulation for a spherical object. More recently, Morgan et al. demonstrate robust finger-gaiting for object reorientation using actor-critic reinforcement learning~\cite{Morgan2021-ny} and multi-modal motion planning~\cite{Morgan2022-xt}, both in conjunction with a compliant, highly underactuated hand designed explicitly for this task. \citet{Bhatt2022-fr} also demonstrate robust finger-gaiting and finger-pivoting manipulation with a soft compliant hand, but these skills were hand-designed and executed in an open-loop fashion rather than autonomously learned.

Model-free RL has also led to significant progress in dexterous manipulation, starting with {OpenAI's} demonstration of finger-gaiting and finger-pivoting~\cite{OpenAI2019-ng} trained in simulation and translated to real hardware. However, this approach uses extensive extrinsic sensing infeasible outside a lab setting, and relies on support surfaces such as the palm underneath the object. \citet{Khandate2022-qt} show dexterous finger-gaiting and finger-pivoting skills using only precision fingertip grasps to enable both palm-up and palm-down operation, but only on a range of simple convex shapes and in a simulated environment. \citet{Makoviychuk2021-ko} showed that GPU physics could be used to accelerate learning skills similar to OpenAI's. \citet{Allshire2021-qp}  used extensive domain randomization and sim-to-real transfer to re-orient a cube but used a table top as an external support surface. \citet{Chen2021-ig, Chen2022-ud} demonstrated in-hand re-orientation for a wide range of objects under palm-up and palm-down orientations of the hand with extrinsic sensing providing dense object feedback. \citet{Sievers2022-lb, Pitz2023-kj} demonstrated in-hand cube reorientation to desired pose with purely tactile feedback. \citet{Qi2022-wy} used rapid motor adaptation to achieve effective sim-to-real transfer of in-hand manipulation skills for small cylindrical and cube-like objects. In our case, the exploration ability of SBP allows learning of policies for more difficult tasks, such as in-hand manipulation of non-convex and large shapes, with only intrinsic sensing. We also achieve successful and robust sim-to-real transfer without extensive domain randomization or domain adaptation by closing the sim-to-real gap via tactile feedback.

\section{Problem Description}
\label{sec:nonhol}

We focus on the problem of achieving dexterous in-hand manipulation while simultaneously securing the manipulated object in a precision grasp. Keeping the object stable in the grasp during manipulation is needed in cases where a support surface is not available or the skill must be performed under different directions for gravity (i.e. palm up or palm down). However, it also creates a difficult class of manipulation problems, combining movement of both the fingers and the object with a constant requirement of maintaining stability.

Formally, our goal is to obtain a policy for issuing finger motor commands to achieve a desired object transformation. The state of our system at time $t$ is denoted by $\bm{x}_t=(\bm{q}_t, \bm{p}_t)$, where $\bm{q} \in \mathcal{R}^d$ is a vector containing the positions of the hand's $d$ degrees of freedom (joints), and $\bm{p} \in \mathcal{R}^6$ contains the position and orientation of the object with respect to the hand. An action (or command) is denoted by the vector $\bm{a} \in \mathcal{R}^d$ comprising new setpoints for the position controllers running at every joint.

For parts of our approach, we assume that a model of the forward dynamics of our environment (\textit{i.e.} a physics simulator) is available for planning or training. We denote this model by $\bm{x}_{t+1} = F(\bm{x}_t, \bm{a}_t)$. We will show however that our results transfer to real robots using standard sim-to-real methods.  

As discussed above, we also require that the hand maintains a stable precision grasp of the manipulated object at all times. Overall, this means that our problem is characterized by a high-dimensional state space, but only small parts of this state space are accessible to us: those where the hand is holding the object in a stable precision grasp. Furthermore, the transition function of our problem is non-holonomic: the subset of fingers that are tasked with holding the object at a specific moment, as well as the object itself, must move in a concerted fashion. Conceptually, the hand-object system must evolve on the complex union of high-dimensional manifolds that form our accessible states. Still, the problem state space must be effectively explored if we are to achieve dexterous manipulation with large object reorientation and finger gaiting.

\section{Sampling-based State Space Exploration}
\label{sec:sbp}

To effectively explore our high-dimensional state space characterized by non-holonomic transitions, we turn to the well-known Rapidly-exploring Random Trees (RRT) algorithm. We leverage our knowledge of the manipulation domain to induce tree growth along the desired manifolds in state space. In particular, we expect two conditions to be met for any state: (1) the hand is in contact with the object only via fingertips, (2) the distribution of these contacts must be such that a stable grasp is possible.

\begin{algorithm}[t]
\caption{General-purpose non-holonomic RRT (G-RRT)}\label{alg:vanilla}
\begin{algorithmic}[1]
\Require Tree containing root node, $G$; $N \gets 1$

\While{$N<N_{max}$} \label{line:mainloop}
\State $\bm{x}_{sample} \gets$ random point in state space
\State $\bm{x}_{node} \gets$ node closest to $\bm{x}_{sample}$ currently in G
\State $d_{min} \gets \infty; \bm{x}_{new} \gets $ NULL
\While{$k<K_{max}$} \label{line:loop}
\State $\bm{a} \gets \mathcal{N}(0, \alpha\bm{I})$ random action
\State $\bm{x}_a \gets F(\bm{x}_{node},\bm{a})$
\If{Stable($\bm{x}_a$) \textbf{and} $dist(\bm{x}_{sample},\bm{x}_a) < d_{min}$ }\label{line:stable}
\State $d_{min} \gets dist(\bm{x}_{sample},\bm{x}_a)$
\State $\bm{x}_{new} \gets \bm{x}_a$
\EndIf
\State $k \gets k+1$
\EndWhile
\If{$\bm{x}_{new}$ is not NULL}
\State Add $\bm{x}_{new}$ to G with $\bm{x}_{node}$ as parent
\State $N\gets N+1$
\EndIf
\EndWhile
\Statex
\Return G
\end{algorithmic}
\end{algorithm}

Assuming system dynamics $F()$ are available and fast to evaluate, we use here the general non-holonomic version of the RRT algorithm, which is able to determine an action that moves the agent towards a desired sample in state space via random sampling. We use the same version of this algorithm as described, for example, by ~\citet{King2016-iv}, which we recapitulate here in Alg.~\ref{alg:vanilla} and refer to as G-RRT.

The essence of this algorithm is the \textbf{while} loop in line~\ref{line:loop}: it is able to grow the tree in a desired direction by sampling a number $K_{max}$ of random actions, then using the transition function $F()$ of our problem to evaluate which of these produces a new node that is as close as possible to a sampled target. 

Our only addition to the general-purpose algorithm is the stability check in line \ref{line:stable}: a new node gets added to the tree only if it passes a stability check. This check consists of advancing the simulation for an additional 2s with no change in the action; if, at the end of this interval, the object has not been dropped (i.e. the height of the object is above a threshold) then the new node is deemed stable and added to the tree. Assuming a typical simulation step of 2 ms, this implies 1000 additional calls to $F()$ for each sample.

Overall, the great advantage of this algorithm lies in its simplicity and generality. The only manipulation-specific component is the aforementioned stability check. However, its performance can be dependent on $K_{max}$ (i.e. number of action samples at each iteration), and each of these samples requires a call to the transition function. This problem can be alleviated by the advent of highly efficient and massively parallel physics engines implementing the transition function, which is an important research direction complementary to our study.

We note that in previous work \cite{Khandate2023-fc} we used an additional condition on the state: the hand must maintain at least three fingers in contact with the object\footnote{Three contacts are the fewest that can achieve stable grasps without relying on torsional friction, which is highly sensitive to the material properties of the objects in contact}. Empirically, we noticed that the three contact constraint had minimal impact on the results of G-RRT. Thus, we discontinued the contact constraint in this work. 

In the same prior work, we also introduced a manipulation-specific version of the non-holonomic RRT algorithm dubbed M-RRT, which does not require the use of the transition function for stability checks. Instead, M-RRT uses manipulation kinematics alone to explore manifolds defined by the contact constraints. This version also required that the hand must maintain at least three fingers in contact with the object. While the M-RRT version has the potential to obtain significant speedups by foregoing dynamic simulation, G-RRT is appealing in its generality and simplicity. As both versions are able to effectively explore the state space, we focus this study on the G-RRT version of our algorithm and plan additional comparisons between these two versions for future work.

\section{Reinforcement Learning}
\label{sec:rl}

While the exploration method we have discussed so far is capable of exploring the complex state space of in-hand manipulation and identifying approximate transitions that follow the complex manifold structure of this space, it does not provide directly usable policies.  The state transitions themselves provide no mechanism to act in states that are not part of the tree, or to act under slightly different transition functions.

To generate closed-loop policies able to handle variability in the encountered states, we turn to RL algorithms. Critically, we rely on the trees generated by our sampling-based algorithms to ensure effective exploration of the state space during policy training.

\begin{algorithm}[t]
\caption{R$\times$R (+ IPT)}\label{alg:rxr+ipt}
\begin{algorithmic}[1]
\Require Tree $G$, randomly initialized actor $\pi_\theta$ and critic $V_\psi$, learning rates $\eta$ and $\nu$
\Statex \texttt{\# Execute Sampling-based Planning}
\State Tree, $G$ $\gets$ G-RRT() \Comment{Alg \ref{alg:vanilla}}
\Statex \texttt{\# Extract top trajectories}
\State Let $\bm{x}_{root}$ be the root state and  $\bm{x}_{goal}$ be the desired goal state towards achieving the manipulation task of interest.
\label{line:rxr_extract_start}
\State $D \gets \{\}, P \gets 1$
\While{$P < P_{max}$}
\State Sample goal $\bm{x}_{goal}$ \Comment{fixed or randomized}
\State $\bm{x}_{node} \gets$ node closest to $\bm{x}_{goal}$ now in G
\State $\tau = \{\bm{x}_{node}\}$ 
\While {$\bm{x}_{node} \neq \bm{x}_{root}$}
\State \textbf{delete}  $\bm{x}_{node}$ from $G$
\State Let $\bm{x}_{parent}$ be parent of $\bm{x}_{node}$
\State $\tau \gets \bm{x}_{parent} \cup \tau$  \Comment{nodes in reverse}
\State $\bm{x}_{node} \gets \bm{x}_{parent}$
\EndWhile
\State $D \gets D \cup \tau$,  $P \gets P + 1$
\EndWhile
\label{line:rxr_extract_end}

\Statex \texttt{\# Imitation Pre-training} \Comment{Optional}
\Statex Get warm-start policy and critic 
\State $\pi_{\theta}$, $V_\psi$ $\gets$ IPT($\mathcal{D}$, $\pi_\theta$, $V_\psi$, $\eta$, $\nu$) \Comment{ from Alg \ref{alg:impt}}
\label{line:rxr_ipt}

\Statex \texttt{\# Reinforcement Learning}
\For{$N$ iterations}
\label{line:rxx_rl_start}
\For{$E$ episodes}
\State Sample  $\bm{x}_{0} \sim \mathcal{D}$ the buffer of best states from sampling-based planning.
\State Collect rollout $\tau$ by following policy $\pi_\theta$ from initial state $\bm{x}_{0}$.
\State $\mathcal{R} \leftarrow \mathcal{R} \cup \tau$
\EndFor
\Statex Update $\pi_{\theta}$ w.r.t policy loss $\mathcal{L}_{\pi}$ on $\mathcal{R}$ with RL algorithm of choice
\State $\theta \leftarrow \theta - \eta\nabla_{\theta} \mathcal{L}_{\pi}(\theta)$ 
\Statex Update critic $V_{\psi}$
\State $\mathcal{L}_{V} = \displaystyle\mathop{\mathbb{E}}_{\bm{o}_t \sim R}\left[(V_\psi(\bm{o}_t)-V^{targ}_t)^2\right]$
\State $\psi \leftarrow \psi - \nu\nabla_{\psi}L_{V}$
\EndFor
\label{line:rxx_rl_end}
\end{algorithmic}
\end{algorithm}

\subsection{Sampling-based Reset Distribution}
\label{sec:RL}

One mechanism we use to transition information from the sampling-based tree to the policy training method is via the reset distribution: we select relevant paths from the planned tree and then use the nodes therein as reset states for policy training.

We note that the sampling-based trees as described here are task-agnostic. Their effectiveness lies in achieving good coverage of the state space (usually within pre-specified limits along each dimension). Once a specific task is prescribed (e.g. via a reward function), we must select states from paths through the tree that are relevant to the task. After selecting said states, we use a uniform distribution over these states as a reset distribution for RL. We describe this concretely in Alg \ref{alg:rxr+ipt}. In particular, lines \ref{line:rxr_extract_start} - \ref{line:rxr_extract_end} show the heuristic approach we use to select paths through the tree to compose the set of reset states. Lines  \ref{line:rxx_rl_start} - \ref{line:rxx_rl_end} show reinforcement learning from using these reset states for exploration. We note that other selection mechanisms are also possible; a promising and more general direction for future studies is to select tree branches that accumulate the highest reward.

This approach has a theoretical grounding in related work showing that derived reset distributions can be used to aid exploration. Let $\rho$ denote the initial state distribution of the MDP to be solved with policy-gradient RL. Recent results~\cite{Agarwal2020-pp} show that it is beneficial to compute policy gradients for policy $\pi$ under a different initial distribution $\mu$ if it enables sufficient exploration. Let $d^{\pi}_{\mu}$ denote the stationary state distribution induced by policy $\pi$ under the initial state distribution $\rho$ and $d^{\pi^*}_{\mu}$ be some stationary distribution of the optimal policy $\pi^*$ initial state distribution $\rho$. Improved exploration can be achieved if $d^{\pi}_{\mu}$ sufficiently covers $d^{\pi^*}_{\rho}$. Here, we obtain such a favorable distribution $\mu$ from plans derived via sampling-based planning.

As this approach is compatible with online RL wherein policy rollouts are collected from a new set of states every episode, both off-policy and on-policy RL are equally feasible. However, we use on-policy learning due to its compatibility with GPU physics simulators and relative training stability.

\subsection{Imitation Pre-training}

\begin{algorithm}[t]
\caption{Imitation Pre-training (IPT)}\label{alg:impt}
\begin{algorithmic}[1]
\Require Sampling-based trajectories (state-only) as demonstration $\mathcal{D}$, randomly initialized policy $\pi$ and value network $V$ with parameters $\theta, \psi$, learning rates $\eta$ and $\nu$
\Statex \texttt{\# Assemble buffer with actions $\mathcal{D}'$}
\For {all consecutive state pairs $\bm{x}_k, \bm{x}_{k+1}$}
\State Compute $\bm{o}_k$ observation vector at $\bm{x}_k$
\State Get $\bm{a}_k$ with Eq \eqref{eq:inv}
\State $\mathcal{D}' \leftarrow \mathcal{D} \cup (\bm{o}_k, \bm{a}_k) $
\EndFor
\Statex \texttt{\# BC with MSE loss $\mathcal{D}'$}
\For{$E_1$ epochs}
\State $L_\pi =\displaystyle \mathop{\mathbb{E}}_{\bm{o},\bm{a} \sim D'} \left[(\bm{a} - \pi_{\theta}(\bm{o}))^2\right]$
\State $\theta \leftarrow \theta - \eta \nabla_{\theta}L_{\pi}(\theta)$
\EndFor
\Statex \texttt{\# Value pre-training}
\For{$E_2$ epochs}
\State Collect rollouts with $\pi_\theta$ and store in $\mathcal{R}$ 
\State $L_{V} = \displaystyle\mathop{\mathbb{E}}_{\bm{o}_t \sim R}\left[(V_\psi(\bm{o}_t)-V^{targ}_t)^2\right]$
\State $\psi \leftarrow \psi - \nu\nabla_{\psi}L_{V}$
\EndFor
\Statex
\Return $\pi_\theta$, $V_\psi$ 
\end{algorithmic}
\end{algorithm}

Relying on sampling-based trajectories exclusively for the reset state distribution disregards potentially valuable information concerning actions and the connectivity between states within exploration trees. Imitation learning, which maps states to actions via supervised learning, can be effective in using such action information embedded within the exploration trees. However, this is not straightforward to execute. The sampling-based trajectories cannot be treated as authentic demonstrations. Instead, they are quasi-static demonstrations composed of a sequence of stable states as a consequence of the stability constraint (Alg \ref{alg:vanilla}, line \ref{line:stable}) we enforce during sampling-based planning. Due to this stability check, the actions are a distorted version of the desired actions under full dynamics. Thus, imitation learning, for instance, with Behavioral Cloning (BC), is unlikely to yield a successful policy due to the approximate nature of demonstrations and inherent challenges arising from distribution shift. Nonetheless, the resulting policies can used as a warm-start policy for reinforcement learning. 

Here we provide a description of this approach. First, we observe that the difference in joint angles between node transitions is a reliable approximation of the desired action, particularly its direction. Augmented by a scaling hyper-parameter, these differences can serve as action labels. For illustration, let $\bm{q}_k$ and $\bm{q}_{k+1}$ be joint angles at successive states. The action labels $\bm{a}_k$ can be obtained by simply 
scaling the difference with a scaling factor $\beta$, i.e $\bm{a}_k = \beta (\bm{q}_{k+1} - \bm{q}_{k})$. In cases where obtaining such action labels is more complex, an inverse dynamics model can be of assistance. Eq \eqref{eq:inv} restates this generally
\begin{align}
\label{eq:inv}
    \bm{a}_k &= g(\bm{x}_{k}, \bm{x}_{k+1})
\end{align}
where $g$ is the inverse function to compute action labels between two successive states of a demonstration.

We then utilize these demonstrations via learning an imitation policy to bootstrap reinforcement learning. To retain the benefits from any generalization properties achieved via imitation learning, we pre-train the critic network on rollouts of the imitation policy to avoid washing it out with a randomly initialized critic \cite{Hansen2022-pc}. This method aligns with contemporary approaches for integrating reinforcement learning with imitation learning which go beyond simply augmenting the online replay buffer with demonstrations \cite{Hu2023-jt}. Our main approach, supplemented with imitation pre-training, is depicted in Alg. \ref{alg:impt}. Lines \ref{line:rxr_ipt} and \ref{line:rxx_rl_start} show imitation pre-training with sampling-based demonstrations and warm-starting reinforcement learning.

\section{Experimental Setup and Tasks}
\label{sec:results}

\vspace{3mm}
\noindent \textbf{Hardware.} We use the robot hand shown in Fig.~\ref{fig:rxr}, consisting of five identical fingers. Each finger comprises a roll joint and two flexion joints for a total of 15 fully actuated position-controlled joints. For the real hardware setup, each joint is powered by a Dynamixel XM430-210T servo motor. The distal link of each finger consists of an optics-based tactile fingertip as introduced by \citet{Piacenza2020-zp}. 

\begin{figure}
    \centering
    \includegraphics[trim=0mm 20mm 0mm 0mm,clip,width=0.3\textwidth]{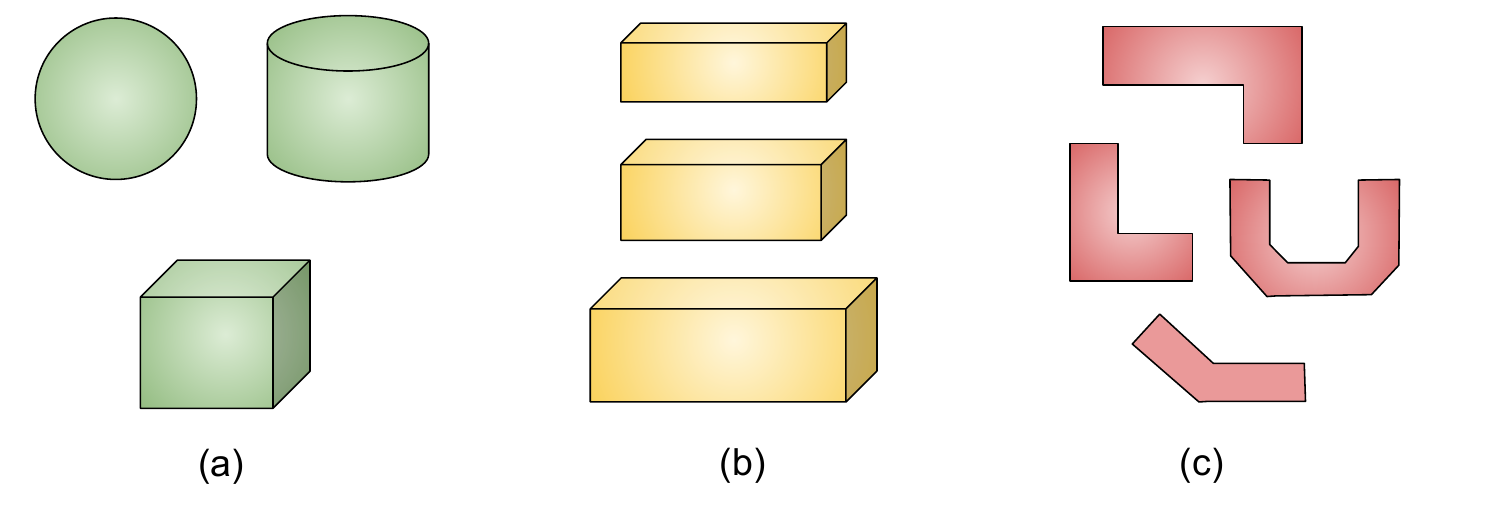}
    \caption{The object shapes for which we learn finger-gaiting. From left to right: the easy, medium and hard categories.}
    \label{fig:objset}
\end{figure}

We test our methods on the object shapes illustrated in Fig.~\ref{fig:objset}. We split this into categories: "easy" objects (sphere, cube, cylinder), "moderate" objects (cuboids with elongated aspect ratios), and "hard" objects (either concave L- or U-shapes). 
We note that in-hand manipulation of the objects in the "hard" category has not been previously demonstrated in the literature.

\vspace{3mm}
\noindent\textbf{Exploration Trees.} The extensive sampling of possible actions, which is the main computational expense of G-RRT (line \ref{line:loop}) lends itself well to parallelization. In practice, we use the IsaacGym \cite{Makoviychuk2021-ko} simulator to parallelize this algorithm at both these levels (16 parallel evaluations of the main loop, and 1024 parallel evaluations of the action sampling loop). 
In general, given the advent of increasingly more powerful parallel architectures for physics simulation, we expect that more general methods that are easier to parallelize might win out in the long term over more problem-specific solutions that are more sample efficient at the individual thread level.

\vspace{3mm} 
\noindent\textbf{Reinforcement Learning}. We train our policies using Asymmetric Actor Critic PPO \cite{Pinto2017-nw, Schulman2017-fz} where we separate actor and critic training for improved performance. All training is done in the IsaacGym simulator and the critic uses object pose $\boldsymbol{p}$, object velocity $\dot{\boldsymbol{p}}$, and net contact force on each fingertip $\boldsymbol{t}_1 \ldots \boldsymbol{t}_{m}$ as feedback in addition to the feedback already as input to the policy network.

For Imitation Pre-training, we compute action labels from state transitions using $\beta=2$ in Eq \eqref{eq:inv}.
We train the critic with rollouts of the imitation policy for 2M steps to mitigate forgetting issues inherent in a randomly initialized critic network. 

\subsection{Tasks}
We evaluate our method on a variety of challenging dexterous in-hand manipulation tasks.
\\~\\
\noindent\textbf{Finger-gaiting}.
First, we focus on the task of achieving large in-hand object rotation about a desired axis. We, as others before~\cite{Qi2022-wy}, believe to this to be representative of this general class of problems, since it requires extensive finger gaiting and object reorientation.

We chose to focus on the case where the only sensing available is hand-centric, either tactile or proprioceptive. Achieving dexterity with only proprioceptive sensing, as biological organisms are clearly capable of, can lead to skills that are robust to occlusion and lighting and can operate in very constrained settings. With this directional goal in mind, the observation available to our policy consists of tactile and proprioceptive data collected by the hand, and no global object pose information. Formally,  the observation vector is
\begin{equation}
\bm{o}_t = [\bm{q}_t, \bm{q}^s_t, \bm{c}_t]    
\end{equation}
where $\bm{q}_t, \bm{q}^s_t \in \mathcal{R}^d$ are the current positions and setpoints of the joints, and $\bm{c}_t \in [0, 1]^m$ is the vector representing binary (contact / no-contact) touch feedback for each of $m$ fingers of the hand.

In this task, where our goal is finger-gaiting for z-axis object rotation, we plan trees where object rotation around the x- and y-axes was restricted to 0.2 radians. Then we select $2 \times 10^4$ nodes from the paths that exhibit the most rotation around the z-axis to construct our reset distribution and for imitation pre-training. On average, each such path comprises 20-30 nodes. We recall that all tree nodes are subjected to an explicit stability check under full system dynamics before being added to the tree; we can thus use each of them as is. 

Similar to our previous work \cite{Khandate2022-qt}, we use a reward function that rewards object angular velocity about the z-axis. In addition, we include penalties for the object's translational velocity and its deviation from the initial position \cite{Qi2022-wy}.
\\~\\
\noindent\textbf{Arbitrary Reorientation}. While finger-gaiting to achieve maximum rotation is a good proof-of-concept task with several applications, we now consider the canonically studied and versatile in-hand manipulation skill of reorientation to a desired pose. We continue with the requirement to achieve in-hand manipulation with only stable fingertip grasps as it enables reorientation in arbitrary orientations of the hand. 

We take on the challenging task of orienting an object from a randomly assigned initial pose to a desired arbitrary orientation, a frequently studied task in dexterous in-hand manipulation \cite{Andrychowicz2020-le, Chen2021-ig}.  In addition to proprioceptive and tactile data collected by the hand, we include current and desired object orientations as inputs to the policy resulting in the following observation vector,
\begin{equation}
\bm{o}_t = [\bm{q}_t, \bm{q}^s_t, \bm{c}_t, \bm{\phi}_t, \bm{\phi}_g]    
\end{equation}
where $\bm{\phi}_t, \bm{\phi}_g \in \mathcal{R}^4$ are the current and desired object orientation respectively. Our reward function is a modification of the one proposed in \cite{Chen2021-ig} as we found the original reward function fails in our setup. Eq. \ref{eq:reward_ar} describes the reward function we use for this task, 

\begin{multline}
    \label{eq:reward_ar}
    r_t = c \cdot \max(\min(\Delta_t - \Delta_{t-1}, \epsilon), -\epsilon) \\ + c_{\text{success}} \cdot \mathbbm{1}[\text{success}]
\end{multline}

where $\Delta_t$ is angular distance between current and desired object orientation, $\epsilon > 0$ is a threshold coefficient, and $c < 0$ is a scaling coefficient. A large positive reward, denoted $c_{\text{success}}$, is added if the agent successfully completes the task. Task completion is achieved through successful reorientation plus satisfaction of heuristics adapted from previous work \cite{Chen2022-ud}. The criteria are described here:
\begin{enumerate}
    \item Reorientation criterion: $\Delta_t < \theta_{\text{thresh}}$
    \item Joint angular velocity criterion: $\|\bm{\dot{q}}_t\|_2 < \dot{q}_{\text{thresh}}$
    \item Object linear velocity criterion: $\|\bm{\dot{x}}_t\|_2 < \dot{x}_{\text{thresh}}$
    \item Object angular velocity criterion: $\|\bm{\omega}_t\|_2 < \omega_{\text{thresh}}$
\end{enumerate}

 Unlike the previous task where a subset of the tree needs to be extracted to aid exploration, here the full tree i.e all nodes of the tree can be used towards assembling the buffer of exploratory reset states. To ensure complete exploration of the state space we use G-RRT trees generated without any constraints on object orientation.
\\~\\
\noindent\textbf{Go-to-root}. While our method improves on sample efficiency for learning policies for arbitrary reorientation, obtaining such policies remains computationally expensive due to the requirement of a few tens of billion simulation training steps. Learning such policies can be overkill; often, in practice, it is sufficient to reorient the object to a fixed canonical pose starting from an arbitrary initial pose. In a packaging line, for example, items arrive at arbitrary orientations. The essential in-hand manipulation task is to reorient objects to a canonical pose before inserting them in the packaging container. 

However, learning to reorient the object to a fixed desired orientation starting from an arbitrary initial pose is still difficult as it suffers the same exploration challenges seen in previous tasks. Fortunately, our method of using exploration trees is naturally well suited for this problem. The robot state with the desired canonical orientation can itself be used as the root node while generating the exploration tree. Thus, we refer to this as the "Go-to-root" task. 

In this task, we aim to learn a policy to reorient the grasped object to reach a desired canonical / root orientation. The inputs to the policy network are similar to the arbitrary reorientation task, consisting of proprioceptive, tactile, and current object pose feedback but excluding the desired goal orientation as it is fixed, resulting in the observation vector in \eqref{eq:g2r_obs}. The reward function and success criteria are identical to the arbitrary reorientation task except with fixed canonical orientation $\bm{\phi}_g$ as the goal.
\begin{equation}
\label{eq:g2r_obs}
\bm{o}_t = [\bm{q}_t, \bm{q}^s_t, \bm{c}_t, \bm{\phi}_t]    
\end{equation}

For imitation pre-training, we extract plans by backtracking from randomly selected nodes with large displacement from the root. In each task, this amounts about about 3K unique observation-action pairs for finger-gaiting and about 15k unique observation-action pairs for go-to-root task.

\subsection{Algorithms and Baselines}
In our experiments, we compare the following approaches:
\\~\\
\noindent\textbf{Ours, R$\times$R}: In this variant, we use the method presented in this paper, relying on exploratory reset states obtained by growing the tree via G-RRT. In all cases, we use a tree comprising $10^5$ nodes as informed the ablation study in Fig~\ref{fig:treesize}.
\\~\\
\noindent\textbf{Ours, R$\times$R + IPT}: In this variant of our method, we use Imitation Pre-training with labels obtained from the grown G-RRT tree, alongside using exploratory reset distribution extracted from the same tree. 
\\~\\
\noindent\textbf{Stable Grasp Sampler (SGS)}: This baseline represents an alternative to the method presented in this paper: we use a reset distribution consisting of stable grasps generated by sampling random joint angles and object orientations. This approach has been demonstrated precision in-hand manipulation with only intrinsic sensing \cite{Khandate2022-qt, Qi2022-wy} for simple shapes. 
\\~\\
\noindent\textbf{Explored Restarts (ER)}: This method selects states explored by the policy itself during random exploration to use as reset states~\cite{Tavakoli2018-ah}. It is highly general, with no manipulation-specific component, and requires no additional step on top of RL training. We implement the "uniform restart" scheme as it was shown to have superior performance on high dimensional continuous control tasks. However, we have found it to be insufficient for the complex state space of our problem: it fails to learn a viable policy even for simple objects.
\\~\\
\noindent\textbf{Fixed Initialization (FI)}: For completeness, we also tried resetting from a single fixed state.
\\~\\
\noindent\textbf{Fixed Initialization (FI + IPT)}: Towards understanding the effect of pre-training, we also tried FI baseline but with a warm start policy obtained from Imitation Pre-training using the G-RRT tree.
\\~\\
\noindent\textbf{Gravity Curriculum (GC)}: Additionally, we evaluated Fixed Initialization with gravity curriculum, linearly annealed from $0 \frac{\text{m}}{\text{s}^2}$ to $-9.81 \frac{\text{m}}{\text{s}^2}$ over the course of 50M timesteps.

\section{Results}

\begin{figure*}
    \centering
    \includegraphics[width=\textwidth]{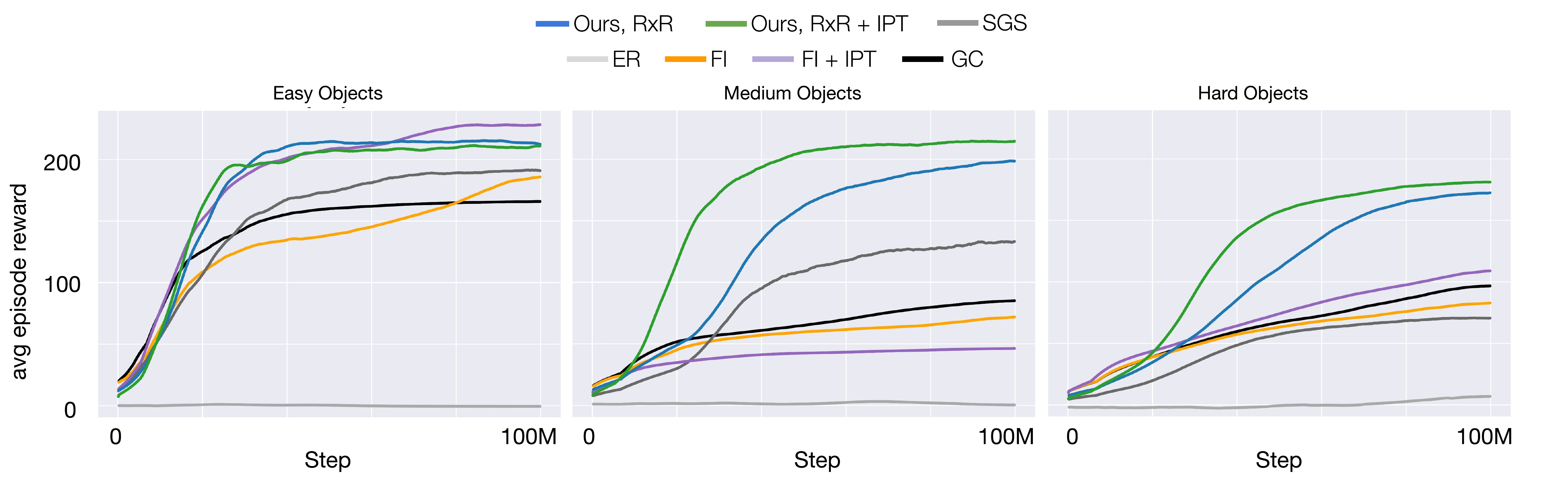}
    \caption{Training performance of our methods and a number of baselines for the Finger-gaiting task on the object categories shown in Fig.~\ref{fig:objset}.}
    \label{fig:results}
\end{figure*}

\begin{figure*}
    \centering
    \includegraphics[width=\textwidth]{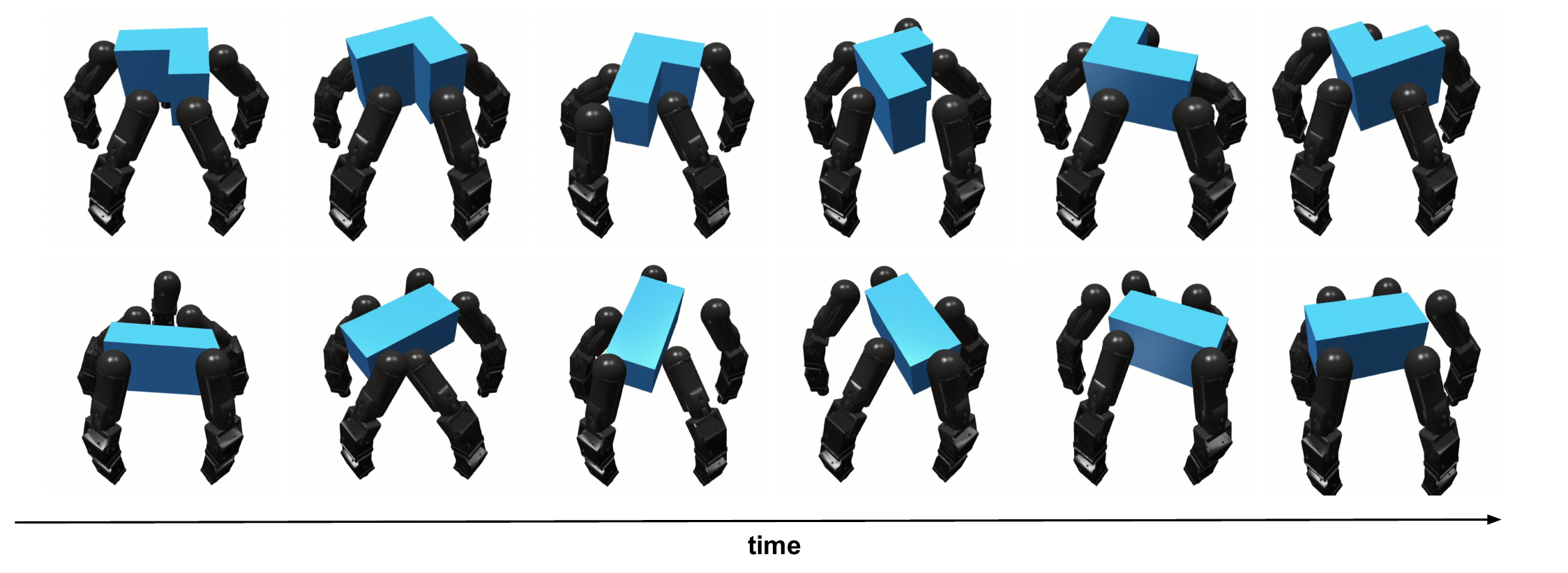}
    \caption{Key frames of the policies for the Finger-gaiting achieved with our method R$\times$R for representative objects in simulation.}
    \label{fig:gaiting}
\end{figure*}

\subsection{Evaluation in simulation}

\label{sec:result_fg}
\noindent\textbf{Finger-gaiting}. Our training results are summarized in Fig.~\ref{fig:results}. We find that our methods (R$\times$R, R$\times$R + IPT) converge within 50M timesteps across all objects for the Finger-gaiting task, consistently outperforming baselines.

\begin{figure*}
    \centering
    \includegraphics[width=\textwidth]{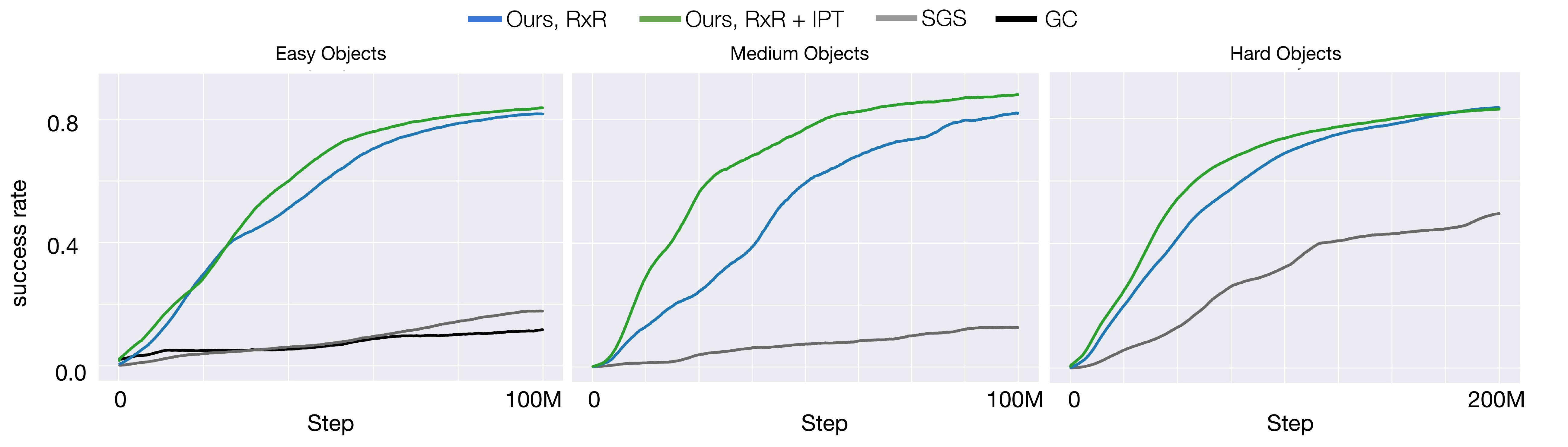}
    \caption{Training performance for the Go-to-root task with our methods and other best-performing baselines.}
    \label{fig:results_g2r}
\end{figure*}
\begin{figure*}
    \centering
    \includegraphics[width=\textwidth]{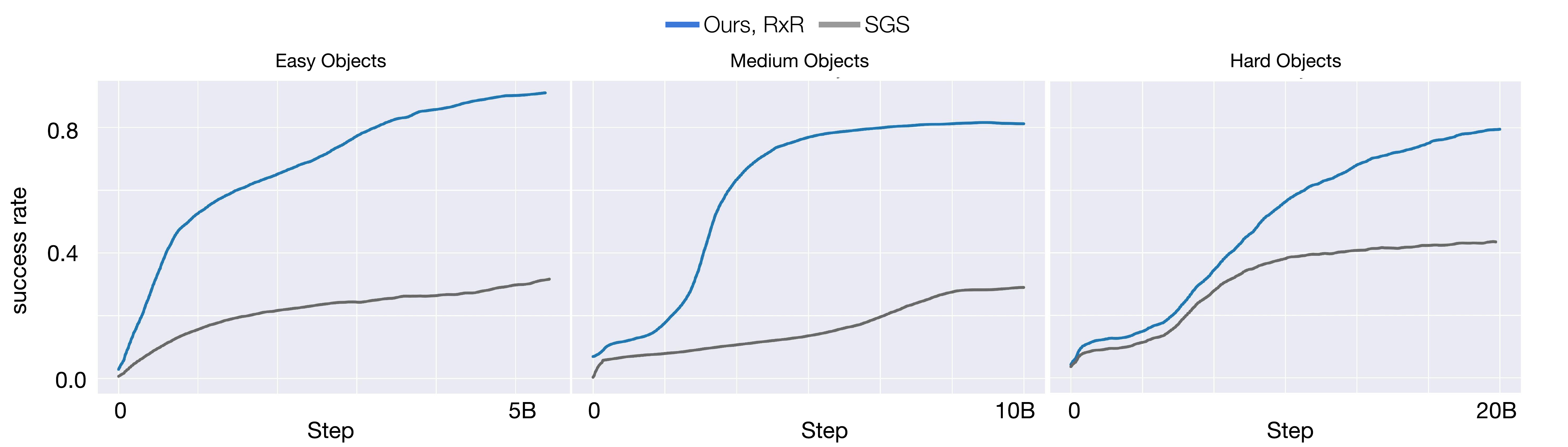}
    \caption{Training performance for the Arbitrary Reorientation task comparing our methods (R$\times$R, R$\times$R + IPT) with Stable Grasp Sampler (SGS) baseline.}
    \label{fig:results_ar}
\end{figure*}
\begin{figure*}
    \centering
    \includegraphics[width=\textwidth]{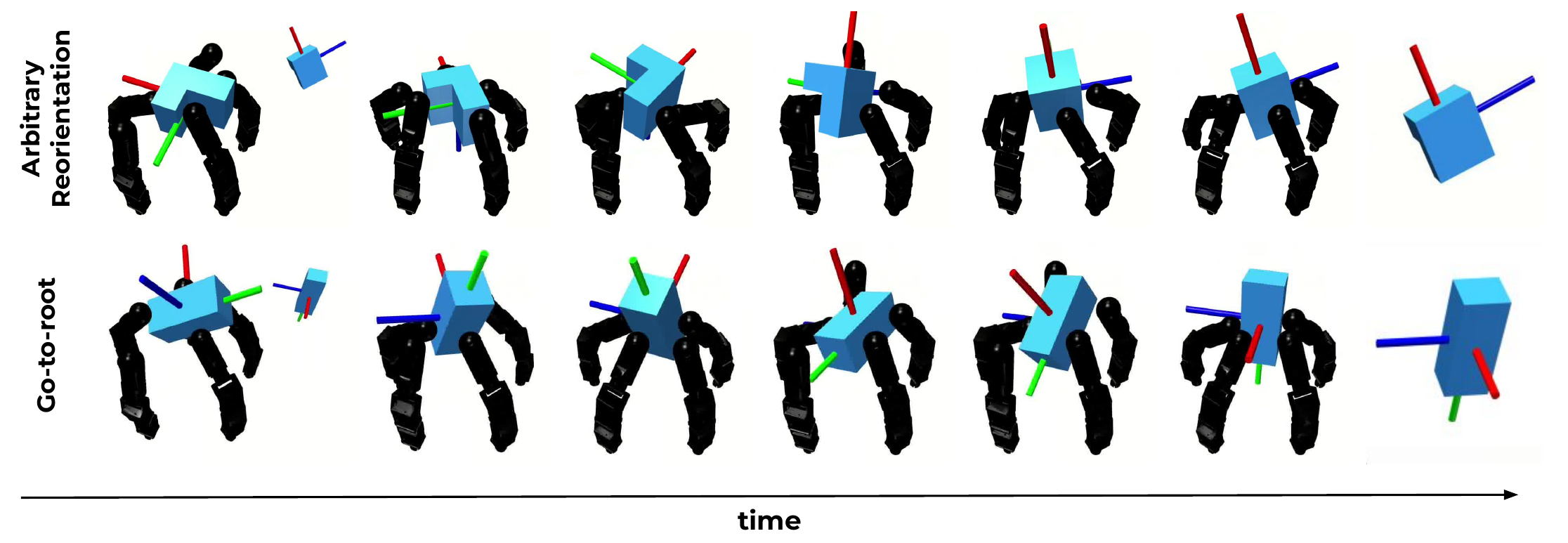}
    \caption{Key frames of goal-reaching tasks, Go-to-root and Arbitrary Reorientation, in simulation.}
    \label{fig:gaiting_ar}
\end{figure*}

On the easy object set, we find that all the methods except the ER baseline learn to gait. However, policies that use a random grasp reset distribution (SGS) and sampling-based reset distribution (R$\times$R) achieve higher performance with robust gaits. We also observe that warm starting with Imitation Pre-training (R$\times$R + IPT) results in faster convergence. Unlike our previous studies, we find that FI also learns finger-gaiting. We attribute this to the improved implementation of our asymmetric actor-critic PPO and increased training steps made possible with the help of GPU physics.

However, as we move to more difficult objects in the medium and hard categories, the performance gap between our methods and baselines broadens. For medium difficulty objects, we find that R$\times$R, R$\times$R + IPT, and SGS again all learn to gait, but the policies learned via our methods, R$\times$R and R$\times$R + IPT, are more effective with higher returns.  Again, we observe that warm starting with Imitation Pre-training (R$\times$R + IPT) results in faster convergence and also achieves improved final performance.

Finally, for the hardest object set, a random grasp-based reset distribution is no longer successful. Only R$\times$R and R$\times$R + IPT converge to stable and consistent gaits. Once again, R$\times$R + IPT improves on R$\times$R with respect to rate of convergence and final performance.

Interestingly, FI + IPT baseline that uses a warm start policy (obtained from Imitation Pre-training using the G-RRT tree) performs similarly to FI baseline, failing on all but easy objects. This suggests that exploratory reset distributions may be necessary to derive benefits from Imitation Pre-training. 

To verify the scalability of the method to train policies for multiple objects, we successfully train object-agnostic policies with our R$\times$R approach. To achieve this we simply use object-specific initial distributions in multi-object training. We found the training difficulty to be limited to the hardest object in the set. Importantly, in Sec \ref{sec:realhandeval}, we also verify our policies can be transferred to the real hand.

Next, we present our results in learning policies to reach a desired object orientation i.e. go-to-root and arbitrary reorientation tasks. Additionally, due to the poor performance of many baselines in the previous task, we continue with the best-performing ones. In particular, we continue with SGS and GC baselines. We drop the ER baseline as it uniformly fails across object classes and also drop the FI baseline as fixed initialization is incompatible with both tasks by definition.  
~
\\~\\
\noindent\textbf{Go-to-root}.
We evaluate R$\times$R and R$\times$R + IPT on the Go-to-root task and compare it with SGS and GC baselines. For this task we augment the Gravity Curriculum baseline to use 20 hand-crafted grasps as the reset distribution. These grasps are constructed to appropriately cover the orientation space of the object. We do this to ensure fair comparison between methods, as learning reorientation from a fixed initial orientation to a fixed target is a significantly easier task. 

Our results are summarized in Fig. \ref{fig:results_g2r}. As shown by the curves, go-to-root is a harder task requiring between 100M-200M steps. However, these results indicate that our methods consistently outperform baseline methods and produce effective control policies for the task. Again we find that R$\times$R + IPT improves on R$\times$R in both convergence and final performance across all object sets. 

~
\\~\\
\noindent\textbf{Arbitrary Reorientation}. We evaluate R$\times$R on the Arbitrary Reorientation task. Our results in Fig. \ref{fig:results_ar} plot the success rate. Note that we considered the hand in "palm-down" orientation as it tends to be the desired hand orientation in many applications. We demonstrate that our method outperforms SGS on all objects. Interestingly, our method significantly outperforms SGS baseline even for objects in the easy category, with increasing performance for objects in the medium and hard categories. The keyframes of our policy executing this task are shown in Fig. \ref{fig:gaiting_ar}. 

Overall, a common thread of all experiments presented here is that our methods (R$\times$R, R$\times$R + IPT) enable learning a range of challenging dexterous manipulation tasks, while none of the domain-agnostic methods (ER, FI, GC) are able to learn in-hand manipulation on objects beyond the easy set. 


\subsection{Evaluation on real hand}

\begin{figure*}
    \centering
    \includegraphics[width=\textwidth]{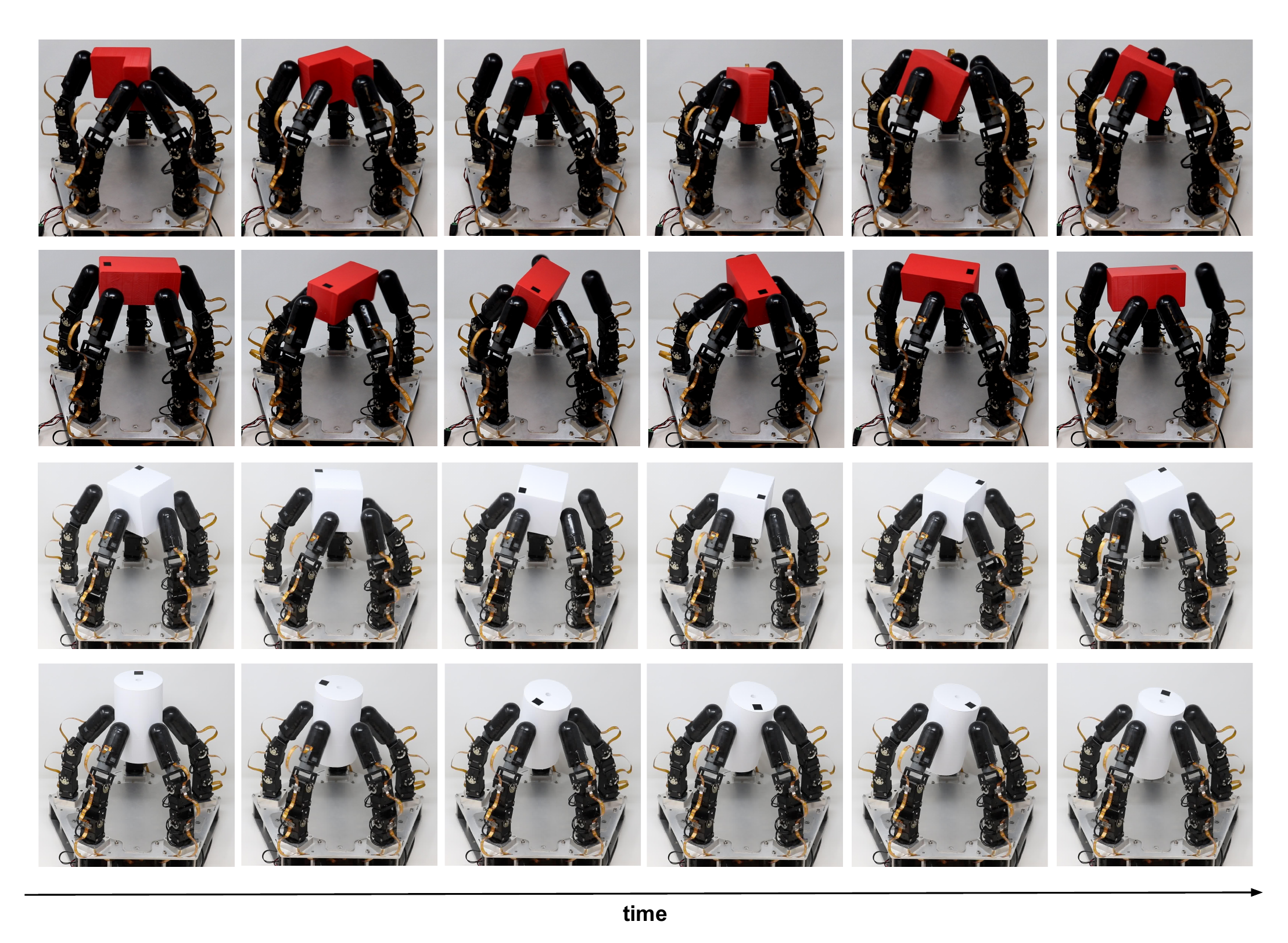}
    \caption{Key frames of the Finger-gaiting policies transferred to the real hand.}
    \label{fig:gaiting_real}
\end{figure*}

\begin{figure*}
    \centering
    \includegraphics[width=\textwidth]{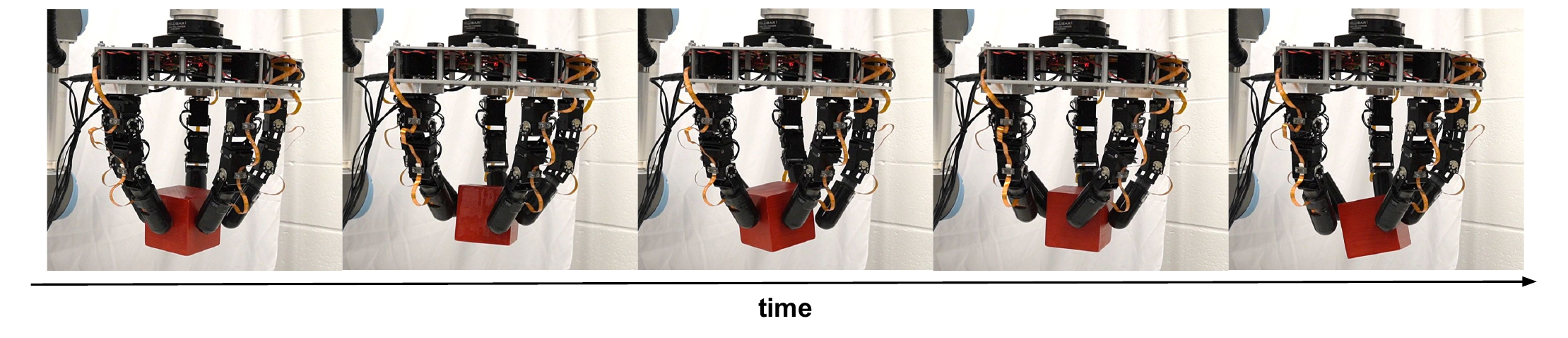}
    \caption{Key frames of a Finger-gaiting policy transferred to the hand in the "palm-down" orientation.}
    \label{fig:gaiting_upside_down_real}
\end{figure*}

\label{sec:realhandeval}
To test the applicability of our method on real hardware, we attempted to transfer the learned Finger-gaiting policies for a subset of representative objects: cylinder, cube, cuboid \& L-shape. We chose these objects to span the range from simpler to more difficult manipulation skills.

To achieve sim-to-real transfer, we take several additional steps. We impose velocity and torque limits in the simulation, mirroring those used on the real motors ($0.6$ rad/s and $0.5$ N$\cdot$m, respectively). We found that our hardware has a significant latency of $0.05$s, which we included in the simulation. In addition, we modified the angular velocity reward to maintain a desired velocity instead of maximizing the object's angular velocity. We also randomize joint origins ($0.1 \text{ rad}$), friction coefficient ($1-40$), and train with perturbation forces (1 N). All these changes are introduced successively via a curriculum. 

For observation, we used the current position and setpoint from the motor controllers with no additional changes. For tactile data, we found that information from our tactile fingers is most reliable for contact forces above 1 N. We thus did not use reported contact data below this threshold and imposed a similar cutoff in simulation. Overall, we believe that a key advantage of exclusively using proprioceptive data is a smaller sim-to-real gap compared to extrinsic sensors such as cameras. We note that an ablation study illustrating the importance of touch feedback is presented in Sec \ref{sec:ablation}.

For the set of representative objects, we ran the respective policy ten consecutive times and counted the number of successful complete object revolutions achieved before a drop. In other words, five revolutions means the policy successfully rotated the object for $1,800^{\circ}$ before dropping it. In addition, we also report the average object rotation speed observed during the trials.

The results of these trials are summarized in Table~\ref{tab:sim2real}. Fig.~\ref{fig:gaiting_real} and Fig.~\ref{fig:gaiting_upside_down_real} show the keyframes of the real hand finger-gaiting we achieved with our policies. Finally, as our finger-gaiting policies do not rely upon vision feedback, our policies are robust to changes in lighting conditions. Examples of our method operating in dynamic lighting conditions can be found in the accompanying video.  

We note that the starting policies used for sim-to-real transfer were trained with trees generated by a version of G-RRT that used the constraint that at least three fingers must be in contact with the object. However, as previously mentioned, we found that this constraint has minimal effect on the trees generated via G-RRT. Nevertheless, we expect similar sim-to-real performance of policies trained using trees generated after forgoing the three contact constraint. These policies are identical to the policies used for sim-to-real transfer, as per visual comparison in simulation. We also note that preliminary attempts to transfer policies for the cylinder and cube also show similar performance.

\begin{table}
    \centering
    \caption{Manipulation performance in real hardware. We report the median number of object rotations achieved before dropping the object in ten consecutive trials, as well as the time needed to perform these rotations.}
    \ra{1.3}
    \begin{tabular}{ccc}
        \midrule
         \phantom{} & Median  &  Mean rotation \\
         \phantom{} &  revolutions & speed (rad/s)\\
         \midrule
         Cylinder & 5 &  0.42 \\
         Cube (s) &  4.5 & 0.44 \\
         Cuboid & 1.5  & 0.44 \\
         L-shape &  1.5 & 0.24 \\
         \midrule
    \end{tabular}
    \label{tab:sim2real}
\end{table}

\section{Ablations}
\label{sec:ablation}

\begin{figure}[t!]
    \centering
    \includegraphics[width=0.45\textwidth]{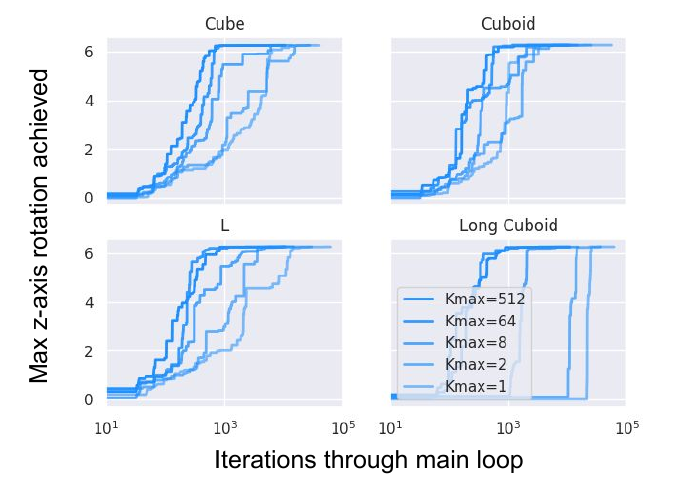}
    \caption{Tree expansion performance for G-RRT. We plot the number of attempted tree expansions (i.e. iterations through the main loop, on a log scale) against the maximum object z-axis rotation achieved by any tree node so far. We plot performance for different values of $K_{max}$, the number of random actions tested at each iteration.}
    \label{fig:rrtcomp}
\end{figure}

\vspace{3mm}
\noindent\textbf{Sampling-based Exploration}. First, we conduct ablation of G-RRT for our object set. We aim to discover how effectively the tree explores its available state space given the number of iterations through the main loop (i.e. the number of attempted tree expansions towards a random sample). As a measure of tree growth, we look at the maximum object rotation achieved around our target axis. We note that any rotation beyond approximately $\pi/4$ radians can not be done in-grasp, and requires finger repositioning. Thus, we compare the maximum achieved object rotation vs. the number of expansions attempted (on a log scale). The results are shown in Fig.~\ref{fig:rrtcomp}. 

As expected, the performance of G-RRT improves with the number $K_{max}$ of actions tested at each iteration. Interestingly, the algorithm performs well even with $K_{max}=1$; this is equivalent to a tree node growing in a completely random direction, without any bias towards the intended sample. However, we note that, at each iteration, the node that grows is the closest to the state-space sample taken at the beginning of the loop. This encourages growth at the periphery of the tree and along the real constraint manifolds, and, as shown here, still leads to effective exploration.

\begin{figure}[]
    \centering
    \includegraphics[width=0.4\textwidth]{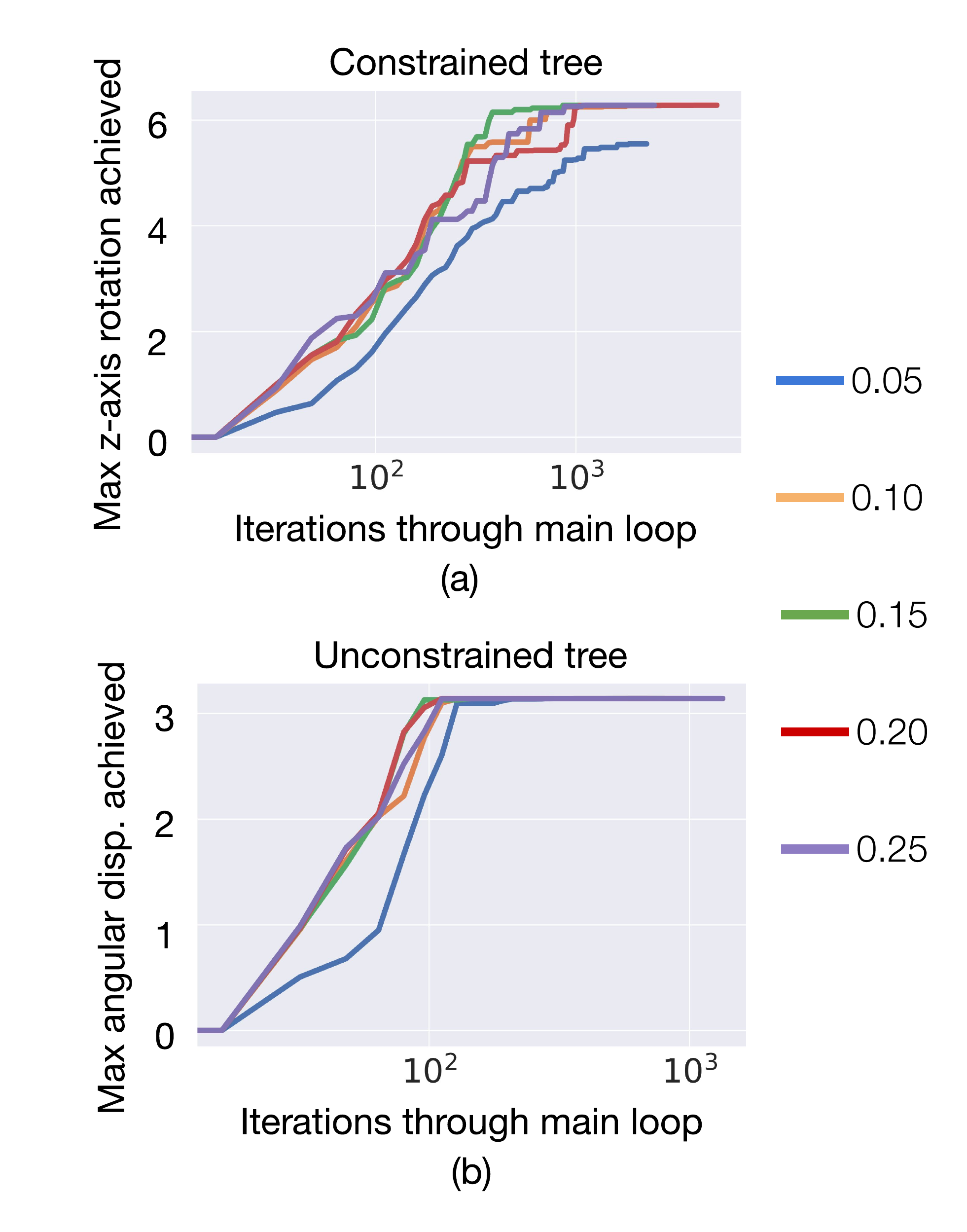}
    \caption{G-RRT action-scale ($\alpha$) ablation. $\alpha = 0.15$ is optimal.}
    \label{fig:grrt_action_abl}
\end{figure}

We also found that G-RRT is sensitive to the action-scale parameter $\alpha$. Fig \ref{fig:grrt_action_abl} compares exploration as measured by angular distance of the farthest node from the root for varying $\alpha$. Among the various settings evaluated, an $\alpha = 0.15$ was the fastest. Interestingly, higher values of action-scale adversely affects the performance of G-RRT. Furthermore, as we will discusses later, not only does $\alpha$ affect the rate of exploration it also impacts the quality of extracted paths. 

\vspace{3mm}
\noindent\textbf{Reinforcement Learning}. We now consider the ablation of training methods. We perform these ablations on the Finger-gaiting and Go-to-root tasks on a non-convex, L-shaped object. 

\begin{figure}[]
    \centering
    \includegraphics[width=0.5\textwidth]{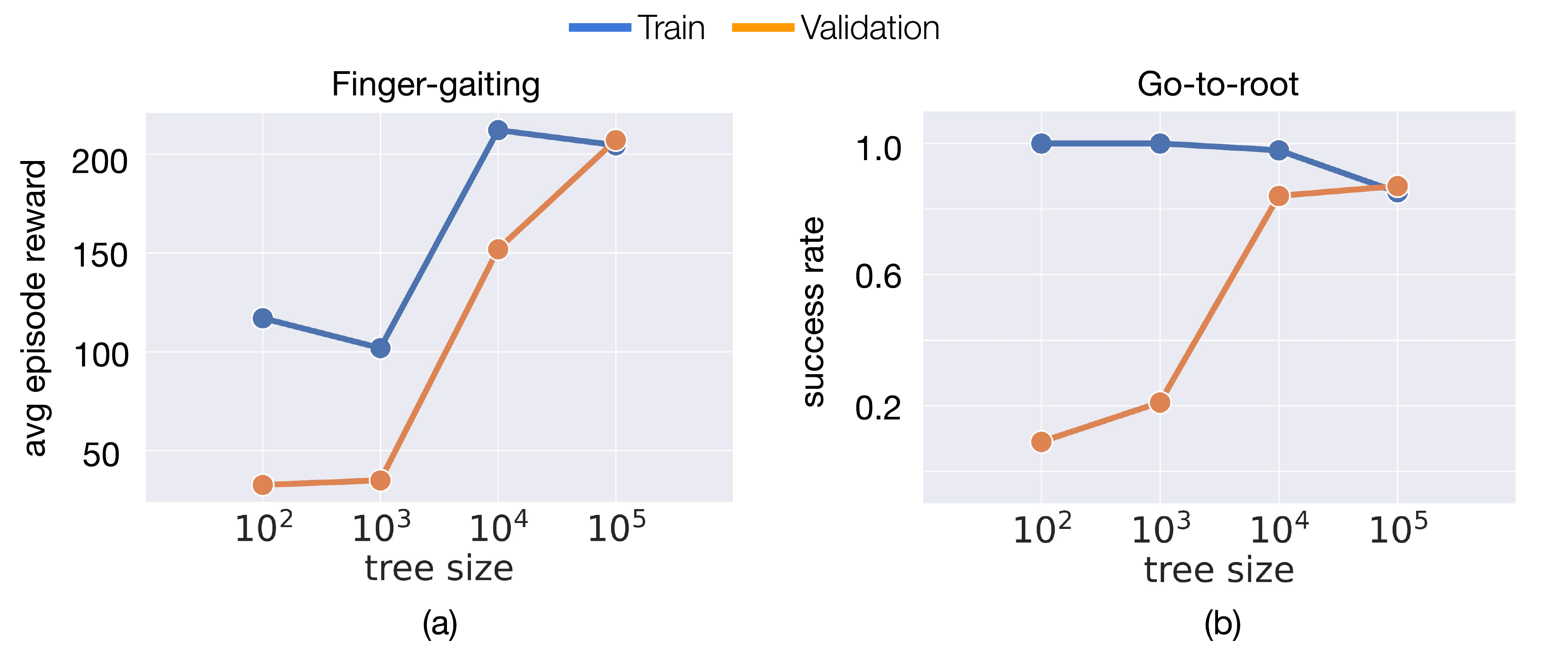}
    \caption{The final performance of policies for Finger-gaiting (left) and Go-to-root (right) tasks using trees of increasing sizes. Validation performance is with reset states from paths extracted from the largest tree.}
    \label{fig:treesize}
\end{figure}

We begin by studying the impact of the size of the tree used in extracting reset states. Fig~\ref{fig:treesize}(a) summarizes our results for learning a Finger-gaiting policy using trees of different sizes grown via G-RRT. Qualitatively, we observe that, as the tree grows larger, the top 100-400 paths sampled from the tree contain increasingly more effective gaits, likely closer to the optimal policy. We see that we need a sufficiently large tree with at least $10^4$ nodes to enable learning. However, training is most reliable with $10^5$ nodes. This suggests a strong correlation between the quality of states used for reset distribution and sample efficiency of learning. 

Fig~\ref{fig:treesize}(b) summarizes the tree-size ablation results for learning the Go-to-root task. For a very small tree, we see near 100\% success rate at training time. This is rather unsurprising as, at these small tree sizes, the nodes are still close to the root orientation resulting in a trivial reorientation task. As we expect, validation performance on paths extracted from the largest tree increases as the size of the tree generating the initial state distribution increases.

\begin{figure}[t]
    \centering
    \includegraphics[width=0.5\textwidth]{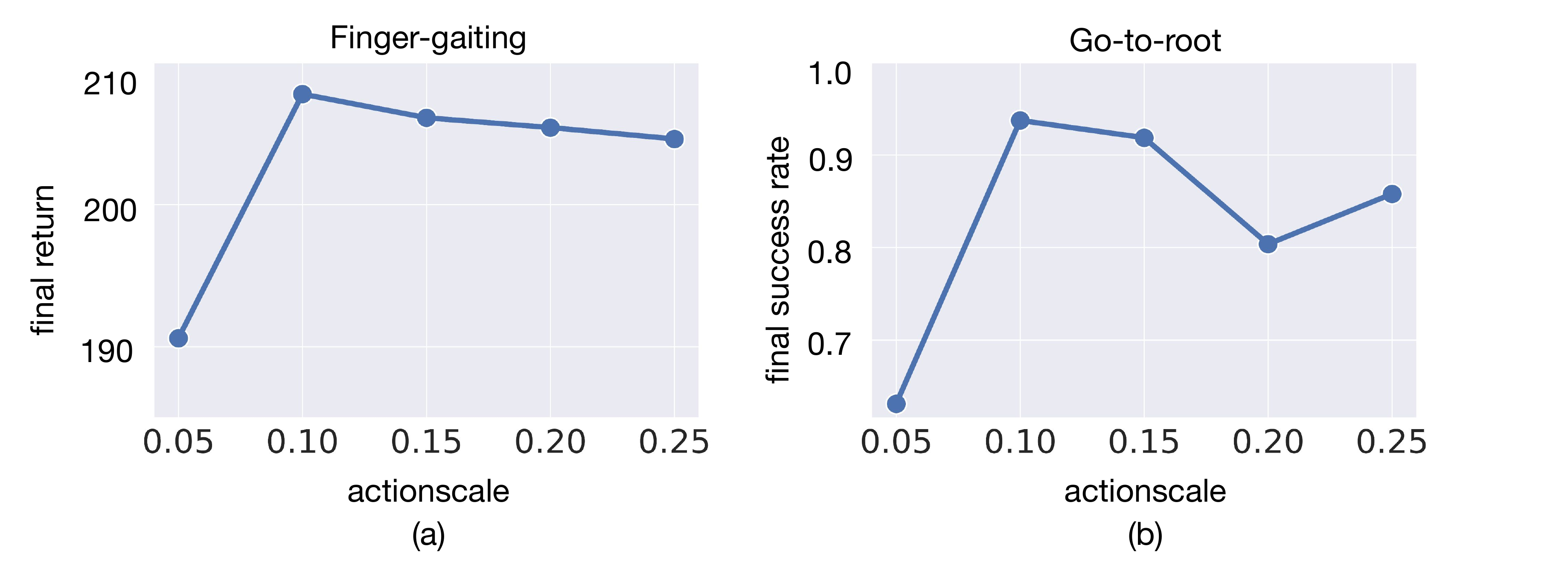}
    \caption{Action scale ablation for the Finger-gaiting and Go-to-root tasks with L.}
    \label{fig:rl_action_abl}
\end{figure}

We also reconsider the action-scale parameter. Besides affecting the rate of exploration it also impacts the quality of paths.  We can infer this by the impact on training performance in Fig \ref{fig:rl_action_abl}. We see a similar trend, as the training performance peaks again around $\alpha = 0.15$ for both Finger-gaiting and Go-to-root. 

\begin{figure}[t]
    \centering
    \includegraphics[width=0.4\textwidth]{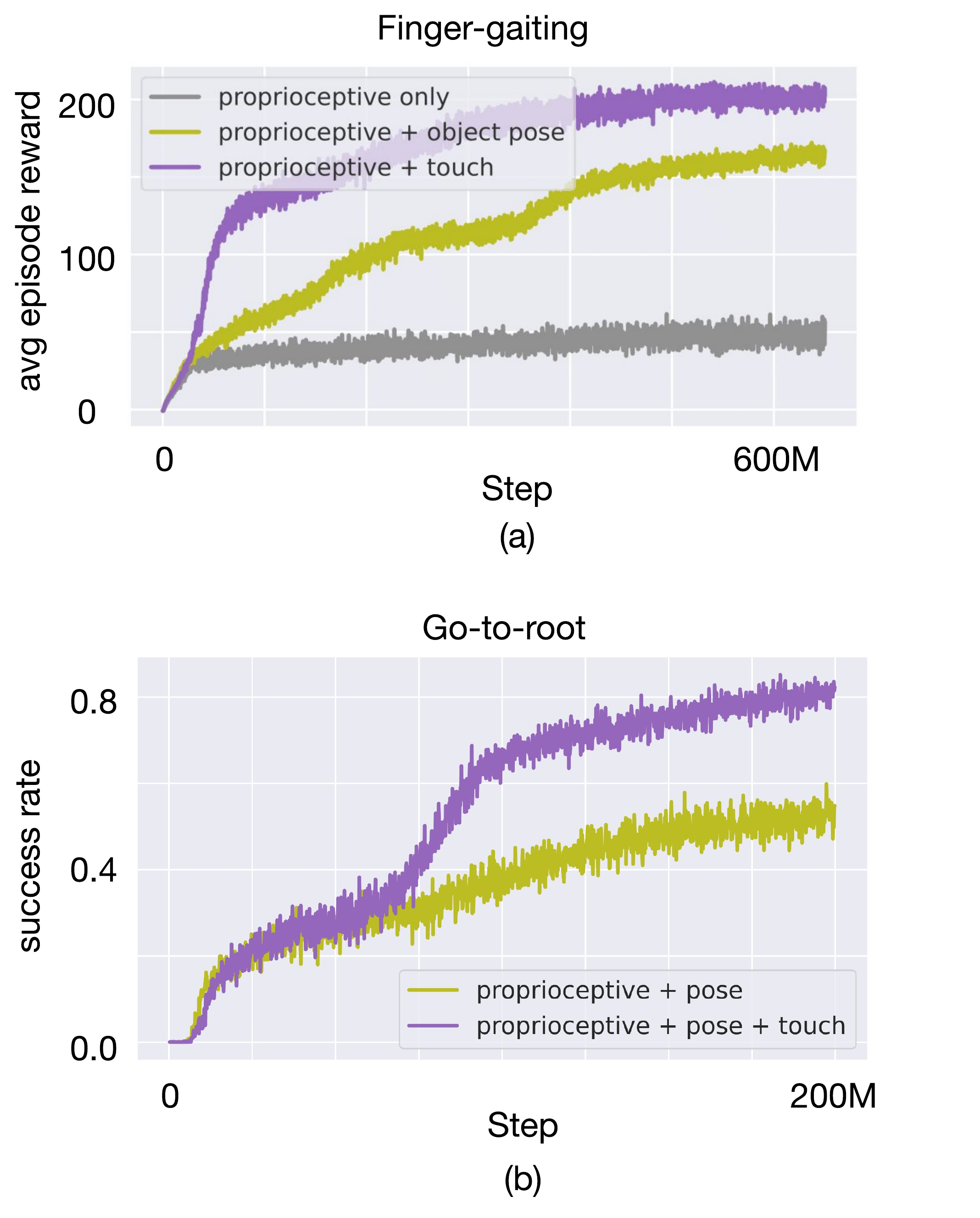}
    \caption{Ablation of policy feedback components which highlights the importance of touch feedback.}
    \label{fig:sense_abltn}
\end{figure}

We also conducted an ablation study of policy feedback. Particularly, we aimed to compare intrinsically available tactile feedback vs. object pose feedback that would require external sensing. Fig.~\ref{fig:sense_abltn} summarizes these results which demonstrate the importance of tactile feedback for both Finger-gaiting and Go-to-root tasks. 

In the Finger-gaiting task, we found that touch feedback is essential in the absence of object pose feedback for all moderate and hard objects. For these objects, we also saw that replacing this tactile feedback with object pose feedback results in slower learning. Similarly, in the Go-to-root task, leaving out touch feedback from policy input also results in slower learning. 

These results underscore the importance of touch feedback for in-hand manipulation skills. Richer tactile feedback such as contact position, normals, and force magnitude can be expected to provide even stronger improvements; we hope to explore this in future work.
\\~\\

\section{Discussion and Conclusions}

The results we have presented show that sampling-based exploration methods make it possible to achieve difficult manipulation tasks via RL. In fact, these popular and widely used classes of algorithms are highly complementary in this case. RL is effective at learning closed-loop control policies that maintain the local stability needed for manipulation, and, thanks to training on a large number of examples, are robust to variations in the encountered states. However, the standard RL exploration techniques (random perturbations in action space) are ineffective in the highly constrained state space with a complex manifold structure of manipulation tasks. Conversely, SBP methods, which rely on a fundamentally different approach to exploration, can effectively discover relevant regions of the state space.

We present multiple methods for conveying exploration information derived through SBP to RL training algorithms. In particular, explored states can be used as an effective reset distribution to enable learning. The transitions between states used during sample-based exploration are also useful surrogates for actions, and can thus be used in an imitation pre-training stage to boost learning. Imitation pre-training serves as an effective strategy to leverage the inherent structure accessible through sampling-based plans. This approach becomes particularly valuable in addressing the challenges of hard motor control tasks within reinforcement learning, not only by facilitating training but also by providing a favorable initialization of actor and critic networks. Recent work in this domain extends the use of the imitation policy beyond mere initialization by demonstrating further benefits when incorporated throughout the training process. This is a promising avenue for achieving further performance gains in reinforcement learning scenarios for hard motor control tasks.

We use our approach to demonstrate a number of dexterous manipulation skills: achieving large in-hand reorientation about a given axis via finger-gaiting and reorientation to a desired object pose, which is either pre-set or randomized. Importantly, we demonstrate finger gaiting precision manipulation of both convex and non-convex objects, using only tactile and proprioceptive sensing. Using only these types of intrinsic sensors makes manipulation skills insensitive to occlusion, illumination, or distractors, and reduces the sim-to-real gap. We take advantage of this by demonstrating our approach both in simulation and on real hardware. We note that, while some applications naturally preclude the use of vision (e.g. extracting an object from a bag), we expect that in many real-life situations, future robotic manipulators will achieve the best performance by combining touch, proprioception, and vision. 

In future work, we believe that our approach can be scaled to tackle even more demanding motor control tasks with numerous degrees of freedom, including both prehensile and non-prehensile manipulation. Bi-manual manipulation seems like a natural candidate application, particularly if it involves coordination with multi-fingered dexterous hands. Furthermore,  sampling-based exploration techniques could be integrated directly into the policy training mechanisms, removing the need for two separate stages during training. We hope to explore all these ideas in the future.
\\~\\

\clearpage
\bibliographystyle{unsrtnat}
\bibliography{paperpile}

\end{document}